\documentclass{article}

\usepackage{microtype}
\usepackage{graphicx}
\usepackage{subfigure}
\usepackage{booktabs} 

\usepackage{hyperref}

\usepackage[accepted]{mlsys2025}

\usepackage{graphicx}
\usepackage{amsmath}
\usepackage[capitalize,noabbrev]{cleveref}
\usepackage{multirow}
\usepackage{tabularx}
\usepackage{tabulary}

\usepackage{booktabs}
\usepackage{multirow}
\usepackage{graphicx} 
\usepackage{xcolor}
\usepackage{amssymb}

\usepackage{tcolorbox}

\definecolor{darkgreen}{rgb}{0.0, 0.5, 0.0}

\mlsystitlerunning{Graph neural networks with configuration cross-attention}

\begin{document}

\twocolumn[
\mlsystitle{Graph neural networks with configuration cross-attention for tensor compilers}



\mlsyssetsymbol{equal}{*}

\begin{mlsysauthorlist}
\mlsysauthor{Dmitrii Khizbullin}{kaust}
\mlsysauthor{Eduardo Rocha de Andrade}{sprout}
\mlsysauthor{Thanh Hau Nguyen}{sprout}
\mlsysauthor{Matheus Pedroza Ferreira}{sprout}
\mlsysauthor{David R. Pugh}{kaust}
\end{mlsysauthorlist}

\mlsysaffiliation{kaust}{King Abdullah University of Science and Technology (KAUST), Thuwal, Saudi Arabia}
\mlsysaffiliation{sprout}{Sprout.ai, London, UK}

\mlsyscorrespondingauthor{Dmitrii Khizbullin}{dmitrii.khizbullin@kaust.edu.sa}

\mlsyskeywords{Machine Learning, MLSys, Tensor compiler, GNN, Attention, TpuGraphs}

\vskip 0.3in

\begin{abstract}
  With the recent popularity of neural networks comes the need for efficient serving of inference workloads. A neural network inference workload can be represented as a computational graph with nodes as operators transforming multidimensional tensors. The tensors can be transposed and/or tiled in a combinatorially large number of ways, some configurations leading to accelerated inference. We propose TGraph, a neural graph architecture that allows screening for fast configurations of the target computational graph, thus representing an artificial intelligence (AI) tensor compiler in contrast to the traditional heuristics-based compilers. The proposed solution improves mean Kendall's $\tau$ across layout collections of TpuGraphs from 29.8\% of the reliable baseline to 67.4\% of TGraph. We estimate the potential CO$_2$ emission reduction associated with our work to be equivalent to over 50\% of the total household emissions in the areas hosting AI-oriented data centers.
\end{abstract}
]



\printAffiliationsAndNotice{}  

\section{Introduction}

Machine learning (ML) continues to gain popularity in solving engineering tasks, including Large Language Models for natural language processing, convolutional and transformer models for computer vision, recommendation models in online services, etc. The majority of the computation associated with ML goes into serving the ML models for inference rather than training them. The need to reduce monetary costs as well as the CO$_2$ footprint of inference workloads leads to significant efforts in the optimization of computations. Typically, ML workloads are launched on specialized accelerators (GPUs, TPUs), which do not provide the same level of on-chip real-time optimization as CPUs do. Consequently, the complexity of optimization of computations for ML accelerators is shifted towards the compiler. Implementation of an enormous quantity of specialized kernels supporting the full matrix formed by a variety of accelerators times a variety of ML models is intangible. One solution to this problem is to employ ML-based tensor compilers. 

\subsection{Related work}

Several attempts have been made to build a highly efficient tensor compiler in recent years. Tensorflow \cite{abadi2016tensorflow} has a rule-based tensor program optimization engine XLA \cite{sabne2020xla} that was studied by \cite{Snider2023OperatorFI}. TVM \cite{chen2018tvm} 
introduces Python-based meta-language to describe the computation and its execution schedule separately, allowing a range of automated optimizations mostly limited to one operator and avoiding operator (kernel) fusion. AutoTVM \cite{chen2018autotvm} introduces optimization of tensor programs based on gradient-boosted trees and TreeGRU and uses the ranking loss for model training rather than element-wise losses like MSE. PyTorch \cite{paszke2019pytorch}, being a framework built with the imperative paradigm in mind, in its recent version, supports TorchScript, a just-in-time (JIT) compiled for the annotated functions and classes. JAX \cite{jax2018github} as a functional meta-language natively supports JIT.

TASO \cite{jia2019taso} performs equivalent graph substitution as a way to fuse kernels. PET \cite{wang2021pet} then builds on top of TASO \cite{jia2019taso} to expand the search space to non-equivalent transformations and apply automatically generated correction kernels. DeepCuts \cite{jung2021deepcuts}, Ansor \cite{zheng2020ansor}, and TensorComp \cite{vasilache2018tensorcomprehensions} rely on heuristics to solve the problem of efficient execution of a computational graph. NN-Meter \cite{nnmeter} presents a latency prediction model based on a combination of heuristics to account for the effects of kernel fusion and a random forest for single-operator latency prediction.

All the aforementioned works mostly rely on heuristics and rules to compile a tensor program. While the compilation time of a heuristics-based algorithm may be very small, it fails to achieve the absolute minimum of program runtime. In this work, we propose an algorithm based on machine learning to optimize a tensor program that is represented as a computational graph. The closest work to ours are \cite{phothilimthana2020learnedmodel} and \cite{xu2023basedon} that use the same dataset and a benchmark TpuGraphs \cite{tpugraphs}. Graph Segment Training (GST) \cite{cao2023graphsegment} uses TpuGraphs as well but reports another metric, OPA, and does not provide a breakdown across the collections.

Apart from TpuGraphs, there are few datasets that represent runtime measurements of computational graphs: Tenset \cite{zheng2021tenset} and the dataset published by the authors of nn-meter \cite{nnmeter}, while none of these explicitly organizes node and edge attributes in a systematic way suitable for machine learning.

\subsection{TpuGraphs dataset and benchmark details}

\begin{figure}[htb]
\vspace{5pt}
\centering
\includegraphics[width=0.98\linewidth]{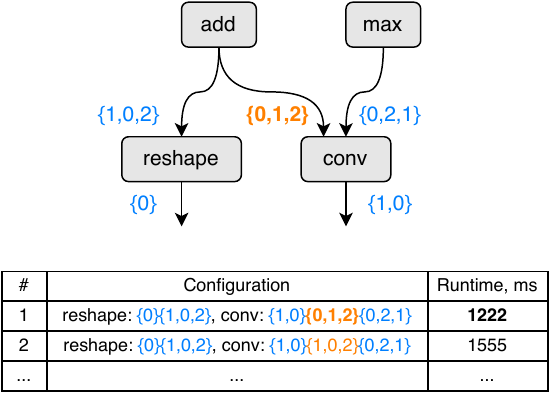}
\caption{
An example of how different tensor layout configurations affect the runtime of the computational (sub-)graph. Configuration 1 is faster than and, consequently, superior to configuration 2.
}
\label{fig:layout_opt}
\end{figure}

\begin{table*}[!htb]
\centering
\caption{The matrix of the 4 \texttt{Layout} collections.}
\label{tab:collectionmatrix}
\begin{tabulary}{\textwidth}{|C|C|C|C|}
\toprule
& & \multicolumn{2}{c}{Configuration sampling strategy} \\
\cmidrule(r){3-4}
& & Random (uniform) & Default (GA-based) \\
\midrule
\multirow{2}{*}{Group of graphs} & XLA (CV, NLP and other) & \texttt{layout-xla-random} & \texttt{layout-xla-default} \\
& NLP (Transformers) & \texttt{layout-nlp-random} & \texttt{layout-nlp-default} \\
\bottomrule
\end{tabulary}
\end{table*}

The only publicly available dataset for the large-scale compiler configuration search is TpuGraphs \cite{tpugraphs}. TpuGraphs contains execution times of an XLA's HLO graph with a specific compiler configuration on a Tensor Processing Unit (TPU v3). TpuGraphs focuses on optimizing tensor layouts and tensor tiling as compiler configurations. Tensor layout optimization dataset comprises 4 collections organized in a matrix shown in \Cref{tab:collectionmatrix}. The two groups of network architectures (\texttt{xla} and \texttt{nlp}) represent two distinct categories of workloads: \texttt{xla} - predominantly computer vision loads, while \texttt{nlp} - exclusively transformer-based natural language processing loads. Each architecture has up to 100'000 different tensor layout configurations and the associated runtimes recorded. The total number of unique architectures in \texttt{layout:xla} collections is 78 with the average number of configurations of over 11,000 (for \texttt{layout:xla::random}), and in \texttt{layout:nlp} collections - 244 with the average number of configurations of over 66,000 (for \texttt{layout:nlp::random}). Another dimension across which the layout dataset is organized is the utilized configuration search strategy: random or genetic-algorithm-based (GA-based, denoted as Default). Even though the final goal is to be able to predict configurations' runtimes, during the dataset creation, some sort of bootstrapping search must be used. Random search gives very wide coverage across all the possible runtimes, whereas the GA-based search focuses more on sampling runtimes in the vicinity of the fastest runtime, making the task of runtime prediction harder and very challenging for the predictive model.

To illustrate the problem of configuration selection we provide an example on Figure \ref{fig:layout_opt}. Here, 4 elementary operations compose a computational graph, while only two of them, reshape and conv, are configurable. A tensor layout can be chosen by the compiler, and the choice results in potentially significantly different runtimes as a result of random or sequential memory access and deep specifics of a particular computational unit. More details can be found in \cite{tpugraphs} Figure 3.

\subsection{Contribution summary}

Our contributions can be summarized as follows:

\begin{itemize}
\item We propose TGraph, a graph neural network (GNN) architecture with cross-channel and cross-configuration attention that achieves state-of-the-art on the TpuGraphs benchmark.
\item We show very efficient training and inference by applying non-configurable node pruning, configuration de-duplication, and compression.
\end{itemize}

\subsection{Societal impact}

We perform a case study to highlight the importance of data center AI workload optimization. According to our estimates the potential impact of this work can be reduction of CO$_2$ emissions equivalent to 50\% (or higher) of household emissions in areas similar to North Virginia, VA. The details can be found in \Cref{sec:casestudy}.

\section{TGraph runtime ranking architecture}

\subsection{Problem specification}

We are looking to find the configuration $\tilde{c}$ that minimizes the tensor program runtime $R(c)$ across the configuration space $C$ for a specific computational graph.

\begin{align}
\tilde{c} = \underset{c \in C}{\text{argmin}} \left( R(c) \right)
\label{eq:problemspec}
\end{align}

As we have only partial knowledge of R(c) in the form of benchmarked data, we are looking for a solution as an approximation $R_{neural}(c)$ of the underlying true $R(c)$.

The configuration space $C$ can be described as $\mathbb{Z}^N$ where $N$ is the number of discrete configurable variables (node and edge attributes) in a specific graph.

\subsection{Data pre-processing}

\subsubsection{Graph pruning}

\begin{figure*}[htb]
\centering
\vspace{8pt}
\includegraphics[width=0.9\textwidth]{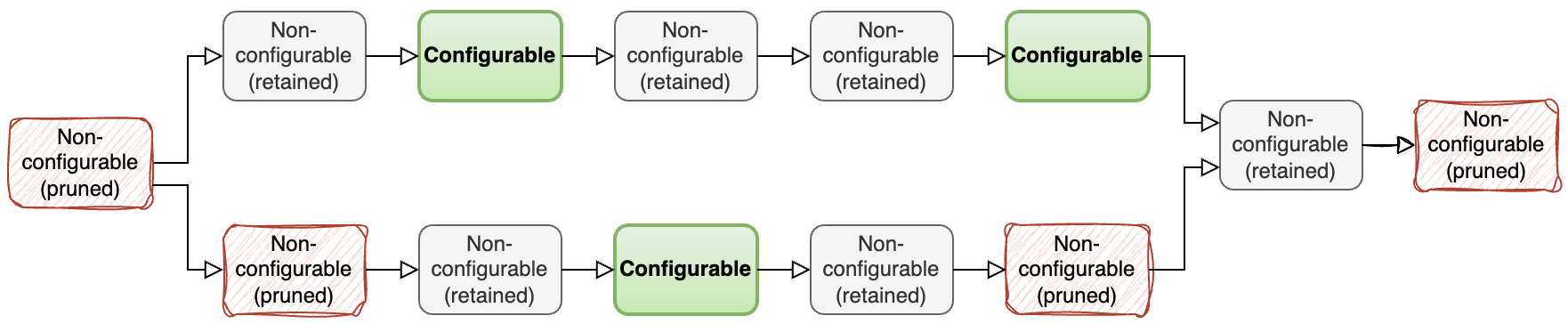}
\vspace{10pt}
\caption{
An example of node pruning. Nodes that are not connected to configurable nodes are removed (red nodes on the diagram). Two disconnected subgraphs are left after pruning.
}
\label{fig:pruning}
\end{figure*}

For layout collections, only Convolution, Dot, and Reshape nodes are configurable. Also, in most cases, the majority of nodes are identical across the configuration set. Thus, we adopt the following pruning strategy: for each graph, we only keep the nodes that are either configurable nodes themselves or are connected to a configurable node, i.e., input or output to a configurable node. By doing this, we transform a single graph into multiple (possibly disconnected) sub-graphs. The possibly disconnected graph does not pose a problem since TGraph has a global graph pooling layer as one of the final layers that fuses the sub-graph information. This way of graph pruning reduces the vRAM usage 4 times and speeds up training by a factor of 5 in some cases. An example of graph pruning is shown on \Cref{fig:pruning}.

\subsubsection{Configuration deduplication}

Most of the configuration sets for layout collections contain a lot of duplication. The runtime for the duplicated configuration sets can vary up to 0.4\% of the mean value. Training on the same configuration sets but different runtime targets makes loss noisy and the training process less stable. Thus, we remove all the duplicated configuration sets for layout collections and leave the smallest runtime value for determinism.

\subsubsection{Lossless configuration compression}

Even with pruning and de-duplication, the RAM usage to load all configurations to the system memory for NLP collections is beyond the RAM capacity. We circumvent that issue by compressing  \texttt{node\_config\_feat} beforehand and only decompressing it on the fly in the data loader after configuration sampling. This allows us to load all data to memory at the beginning of training, which reduces IO/CPU bottlenecks considerably and allows us to train faster. The compression is implemented based on the fact that each \texttt{node\_config\_feat} 6-dim vector (input, output, and kernel) can only have 7 possible values (-1, 0, 1, 2, 3, 4, 5) and, thus, can be represented by a single integer in base-7 (from 0 to $7^6-1$).

\subsubsection{Changing the pad value in \texttt{node\_feat}}
The features in \texttt{node\_feat} are 0-padded. Whilst this is not a problem for most features, for others like \texttt{layout\_minor\_to\_major\_*}, this can be ambiguous since 0 is a valid axis index. Also, the \texttt{node\_config\_feat} are -1 padded, which makes it incompatible with \texttt{layout\_minor\_to\_major\_*} from \texttt{node\_feat}. With that in mind, we re-generate \texttt{node\_feat} with -1 padded, and this allows us to use a single embedding matrix for both \texttt{node\_feat[134:]} and \texttt{node\_config\_feat}.

\subsubsection{Data normalization, embedding and batching}

For \texttt{layout}, the node features are formed as a 140-dimensional vector \texttt{node\_feat} that represents various fields in an XLA’s HLO instruction (a node in an HLO graph) either as they are, or as categorical values using one-hot encoding. We split \texttt{node\_feat} into \texttt{node\_feat[:134]} containing numerical and one-hot-encoded values and \texttt{node\_feat[134:]} that contains the tensor index permutation of the output tensor layout (\texttt{layout\_minor\_to\_major\_*}). The former is normalized to element-wise 0-mean and unit standard deviation (\texttt{StandardScaler} on \Cref{fig:architecture_diagram}), while the latter, along with \texttt{node\_config\_feat}, is fed into a learned embedding matrix (4 channels). We find that the normalization is essential since \texttt{node\_feat} has features like \texttt{*\_sum} and \texttt{*\_product} that can be very high in values compared to the rest of the features and, consequently, disrupt the optimization. Further, we find that the natural way to encode the permutation vectors is to embed them into a low-dimensional vector. For \texttt{node\_opcode}, we also use a separate embedding layer with 16 channels. The input to the network is the concatenation of all aforementioned features. For each graph, we sample on the fly a batch of 64 (for \texttt{default} collections) or 128 (for \texttt{random} collections) configurations to form the input batch. For tile, on the other hand, we opt to use late fusion to integrate \texttt{config\_feat} into the network.

\subsection{Architecture details}

Following the reasoning laid out by \cite{phothilimthana2020learnedmodel}, we employ GraphSAGE \cite{graphsage} as a basis of a graph convolutional block. GraphSage operation can be expressed as 

\begin{align}
S_i^k(\varepsilon) = N_{L2} \left( f_2^k \left( \text{concat}\left(\varepsilon_i, \sum_{j \in \text{neighbors}(i)} f_1^k \left( \varepsilon_j \right) \right) \right) \right)
\label{eq:graphsage}
\end{align}
\vspace{8pt}
where $i$ is the index of a node, $k$ is the index of the layer, $f_{1...2}^k$ - feedforward layers at the specific depth $k$, $N_{L2}$ - $L_2$ normalization, $neighbours(i)$ - a set of immediate neighbours of node $i$.

We construct the graph convolutional block that can be expressed in the following way.

\begin{align}
B_i^k(\varepsilon) = \varepsilon + a \left( \text{concat} \left( \eta_i, A_{cross}(\eta_i) \right) \right)
\label{eq:convblock}
\end{align}

where $a$ is GELU activation, $A_{cross}$ - configuration cross-attention operation, and $\eta_i(\varepsilon)$ is expressed as:

\begin{align}
\eta_i(\varepsilon) = A_{self} \left( S_i^k \left( N_{instance} \left( \varepsilon \right) \right) \right)
\label{eq:convblock_common}
\end{align}

Here $A_{self}$ is the self-attention operation described below, $N_{instance}$ is instance normalization.

\subsubsection{Channel-wise self-attention}

Inspired by the idea of Squeeze-and-Excitation \cite{hu2018squeezeexcite}, we add a channel-wise self-attention layer as a part of the graph convolutional block. We first apply a Linear layer to bottleneck the channel dimensions (8x reduction), followed by ReLU. Then, we apply a second linear layer to increase the channels again to the original value, followed by sigmoid. We finish by applying element-wise multiplication to the obtained feature map and the original input. The idea behind channel-wise self-attention is to capture the correlations between channels and use them to suppress less useful ones while enhancing the important ones.

\begin{align}
A_{self}(\varepsilon) = \varepsilon \circ \sigma \left( f_{squeeze} \left( \text{ReLU} \left( f_{excitation} (\varepsilon) \right) \right) \right)
\label{eq:selfattention}
\end{align}

Here $\circ$ denotes element-wise multiplication.

\subsubsection{Cross-configuration attention}

Another dimension in which we apply the attention mechanism is the batch dimension: across the sampled configurations. We design the cross-configuration attention block that allows the model to explicitly compare each configuration against the others throughout the network. We find this method to be much superior to letting the model infer for each configuration individually and only compare them implicitly via the loss function (\texttt{PairwiseHingeLoss} in this paper). The cross-configuration attention expression comes as follows:

\begin{align}
A_{cross}(\varepsilon) = \varepsilon_i^b \circ \underset{b}{\text{Softmax}} \left( \varepsilon_i^b / T \right)
\label{eq:crossconfigattention}
\end{align}

Here $i$ is the node index, $b$ is the configuration index across the batch dimension, $T$ is a learnable temperature parameter.

By applying the cross-configuration attention layer after the channel-wise self-attention at every block of the network, we observe a significant improvement of the target metric (Kendall's $\tau$), especially for \texttt{default} collections.

\subsubsection{Entire architecture}
\begin{figure*}[htb]
\centering
\vspace{8pt}
\includegraphics[width=0.98\textwidth]{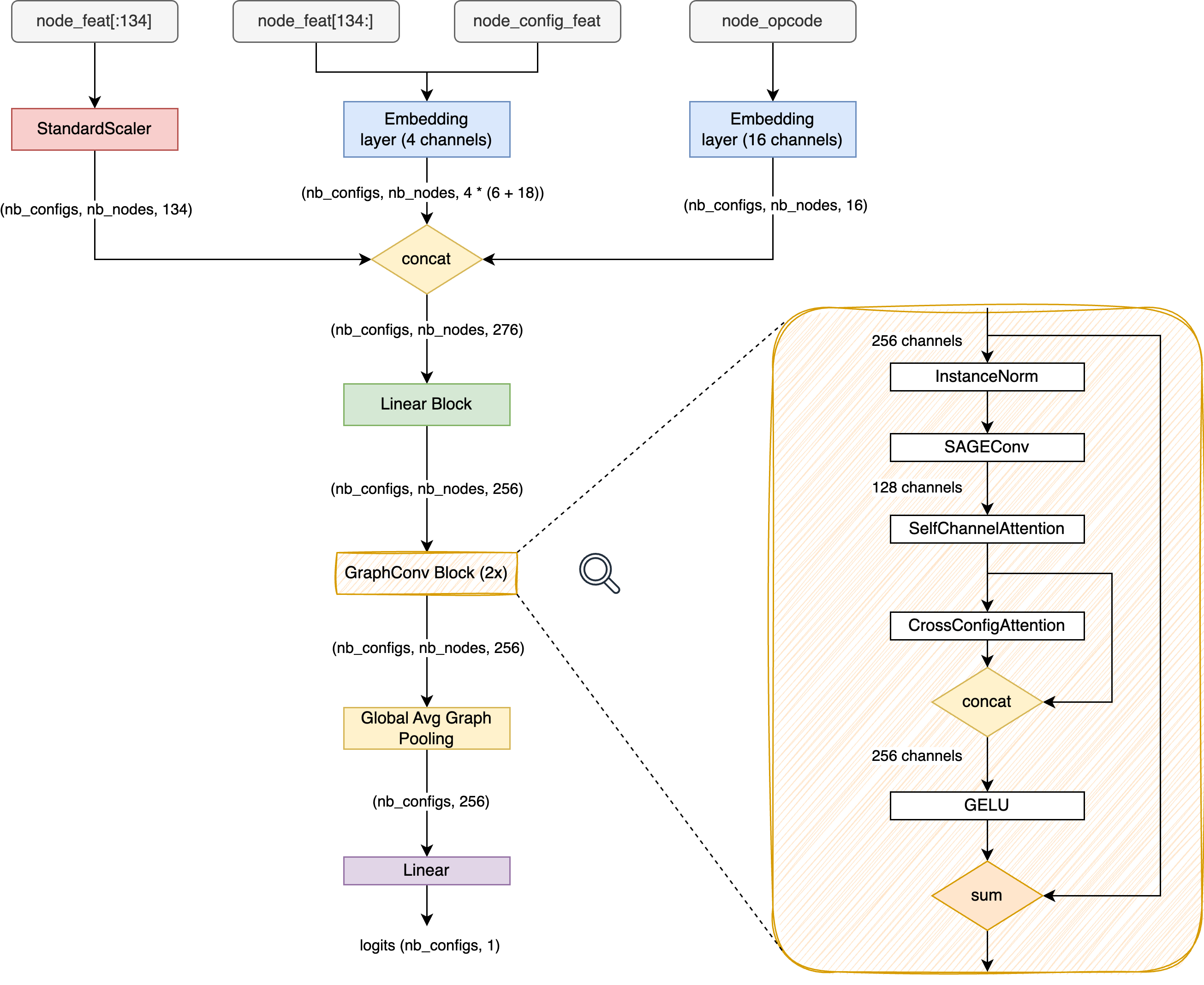}
\vspace{10pt}
\caption{
Architecture diagram of TGraph. $n_{configs}$ is the number of configurations sampled into a batch. $n_{nodes}$ is the number of nodes in the sampled graph after pruning.
}
\label{fig:architecture_diagram}
\end{figure*}

The full architecture of TGraph is shown in \Cref{fig:architecture_diagram}. After feature concatenation, we apply a fully-connected layer, then we apply a stack of 2 graph convolutional blocks $B_i^k$, $k \in 1..2$, then we perform global average pooling over the node dimension indexed by $i$, and finally, we apply another linear layer to eliminate the feature dimension and get the vector of scores $s_c$ where $c$ is the index across the configuration dimension. 

The entire network prediction can be expressed as:

\begin{align}
R_{neural}(X) = f_{out} \left( Pool_{global} \left( B_2 \left( B_1 \left( f_{in} \left( X \right) \right) \right) \right) \right)
\label{eq:fullnet}
\end{align}

where $X$ is the input feature vector, $f_{in}$ - a 2-layer MLP with \{256, 256\} features and GELU activation, $f_{out}$ - linear layer with a single feature and no activation, $Pool_{global}$ - global average pooling across nodes.

\subsection{Training and inference procedures}
\label{sec:training}

\subsubsection{Loss function}

We use the Pairwise Hinge Loss (\texttt{PairwiseHingeLoss}, \cite{thorsen2002clickthrough}, \cite{agarwal2019conterfactual}) loss function for training the model.

\begin{align}
\mathcal{L}(\{r\}, \{s\}) =
  \sum_i \sum_j I[r_i > r_j] \max(0, 1 - (s_i - s_j))
\label{eq:pairwisehingeloss}
\end{align}

where $r_i$ - are the ground truth runtimes, $s_i$ - are the scores predicted by the model.

It is important that the predicted scores $s_i=R_{neural}(c_i)$ do not correspond to the absolute values of runtimes $r_i=R(c_i)$. The applied loss function is a ranking loss function. It trains the model to order (rank) the predicted values in the same way as they are ordered by $R(c)$. The correct ordering is enough to satisfy \Cref{eq:problemspec}.

\subsubsection{Training details}

We train separate model instances for all collections. We've identified that separate models perform better than a joint model trained on all collections or models that were trained on all-\texttt{xla} or all-\texttt{nlp} combinations as well as all-\texttt{random} or all-\texttt{default}.

We use Adam \cite{kingma2014adam} optimizer (specifically AdamW version) with the learning rate of 1e-3, 0.05 of the total number of epochs as linear warm-up, a single-cycle (lifted cosine) learning rate schedule, and weight decay of 1e-5 for non-bias parameters. We apply gradient norm clipping at value 1.0.

We train the \texttt{tile-xla} collection for 17.5 epochs, whereas \texttt{layout-nlp} collections for 1000 epochs and \texttt{layout-xla} collections for 750 epochs.

Training wall-clock time is 2.5 hours per fold per collection measured on RTX4090 with 24 GB RAM. Training one set of models for all collections produces 13.45 kg CO$_2$ as per \cite{Lacoste2019QuantifyingTC}.

\subsubsection{Data splits}
\label{sec:datasplits}

Whereas the official training/validation split is reasonably designed, we, however, employ K-fold cross-validation with $K=20$ on the merged train/validation data splits. We train the first 5 folds to limit the training compute. We then pick the top-4 folds by the validation score to combat the instability of training. This choice comes from the slight instability of training: in rare cases, the training process for a specific fold may get stuck at a local minimum or experience partial parameter corruption due to gradient explosion. In addition, we choose not to split configurations of the same graph into train/validation since it would introduce a train-to-validation leak due to the very high correlation of configuration runtimes within the same graph.

\subsection{Benchmark results}

\begin{table*}[!htb]
  \caption{Experimental results.}
  \label{sample-table}
  \centering
  \begin{tabular}{llp{2.5cm}ll}
    \toprule
    & & \multicolumn{3}{c}{Validation score} \\
    \cmidrule(r){3-5}
    Collection & Metric & TpuGraphs \cite{tpugraphs} & \cite{xu2023basedon} & TGraph (ours) \\
    \midrule
    layout:xla:random  & Kendall's $\tau$ & 0.19 & 0.5285 & \textbf{0.6840} $\pm$ 0.0110 \\
    layout:xla:default & Kendall's $\tau$ & 0.12 & \textbf{0.5887} & 0.4785 $\pm$ 0.0031 \\
    layout:nlp:random  & Kendall's $\tau$ & 0.58 & 0.8387 & \textbf{0.9713} $\pm$ 0.0008 \\
    layout:nlp:default & Kendall's $\tau$ & 0.30 & 0.4841 & \textbf{0.5628} $\pm$ 0.0027 \\
    \midrule
    mean across layout  & Kendall's $\tau$ & 0.298 & 0.610 & \textbf{0.674} \\
    \midrule
    tile:xla           & $M_{tile}$     & -    & 0.8622 & \textbf{0.9694} $\pm$ 0.0021 \  \\
    \bottomrule
  \end{tabular}
\label{tab:evaluation}
\end{table*}

\begin{table*}[htb!]
    \centering
    \caption{Ablation Study.}
    \label{tab:ablation_studies}
    \begin{tabular}{lllll}
        \toprule
        \multirow{2}{*}{Configuration} & \multicolumn{3}{c}{Validation score, Kendall's $\tau$} \\
        \cmidrule(lr){2-5}
        & layout:xla:random & layout:xla:default & layout:nlp:random & layout:nlp:default \\
        \midrule
        Final, all features             & \textbf{0.6840}       & 0.4785           & \textbf{0.9713}       & \textbf{0.5628} \\
        - Channel-wise self-attention   & 0.6737 (\textcolor{red}{-0.0103}) & \textbf{0.4787} (\textcolor{darkgreen}{+0.0002}) & 0.9680 (\textcolor{red}{-0.0033}) & 0.5555 (\textcolor{red}{-0.0073}) \\
        - Cross-configuration attention & 0.6539 (\textcolor{red}{-0.0301}) & 0.4518 (\textcolor{red}{-0.0267}) & 0.9387 (\textcolor{red}{-0.0326}) & 0.5436 (\textcolor{red}{-0.0192}) \\
        - Graph edges                   & 0.5022 (\textcolor{red}{-0.1818}) & 0.3631 (\textcolor{red}{-0.1154}) & 0.7751 (\textcolor{red}{-0.1962}) & 0.3349 (\textcolor{red}{-0.2279}) \\
        \bottomrule
    \end{tabular}
\end{table*}

\subsubsection{Evaluation splits}

TpuGraphs \cite{tpugraphs} dataset does not provide public test data annotations. Hence, we report the cross-validation score according to the \Cref{sec:datasplits}.

\subsubsection{Evaluation metrics}

Kendall's $\tau$ (Kendall's rank correlation coefficient) is used as the metric for \texttt{layout} collections:

\begin{align}
\tau = \frac{2}{n(n-1)} \sum_{i < j} \text{sgn}(s_i - s_j) \text{sgn}(r_i - r_j)
\label{eq:kendalltau}
\end{align}

where $s$ are the predicted scores, $r$ are the ground truth runtimes, $n$ is the batch size.

For the \texttt{tile} collection, the metric is set as:

\begin{align}
M_{\text{tile}} &= 1 - \left( \frac{\text{Best runtime of top-}k\ \text{predictions}}{\text{Best runtime of all configurations}} - 1 \right) \nonumber \\
&= 2 - \frac{\min_{i \in K} r_i}{\min_{i \in A} r_i}
\label{eq:tilemetric}
\end{align}

where $K=5$.

\subsubsection{Details of the inference mode}
\label{sec:inference}

For inference, we use the batch size of 128. However, since the prediction depends on the batch, we leverage the batch further by applying test-time augmentation (TTA) to generate N (10) permutations of the configurations and average the result after sorting it back to the original order. We average the scores of models trained on different folds.

The single-batch wall clock time is 60 ms on average for 1 fold and 240 ms on average for all 4 folds per collection.

\subsubsection{Experimental results}

Our experimental results are summarized in the \Cref{tab:evaluation}. The confidence ranges are reported as 1-sigma. We demonstrate state-of-the-art performance in 4 out of 5 collections. On \texttt{xla-default} Xu 2021 \cite{xu2023basedon} show better results than our work; however, their results may contain an error since \texttt{xla-default} collection is harder than \texttt{xla-random} due to closer and harder-to-distinguish runtime annotations (the pattern is also followed by the results of TpuGraphs \cite{tpugraphs}), but the score of \cite{xu2023basedon} for \texttt{xla-default} is higher than for \texttt{xla-random} which is very implausible.

\subsubsection{Ablation study}

Ablations for channel-wise self-attention, cross-configuration attention, and edges in the graph are collected in Table \ref{tab:ablation_studies}. While the effect of channel-wise self-attention is less obvious but nevertheless noticeable, the effect of cross-configuration attention is substantial, implying that the task of comparing the configurations between each other is easier than predicting the absolute values of runtimes. Additionally, we ablate the edges of the GraphSage GNN to demonstrate how essential the connectivity between the computational nodes is. In tensor compilers the adjacent operators are often fused into a single optimized operator, the procedure commonly know as kernel fusion. For a model solving the problem of predicting computational graph runtimes it is paramount to implicitly learn the "rules" of kernel fusion from data since the early stages of tensor compilation including kernel fusion are treated as a black box.

\subsection{Environmental impact case study}
\label{sec:casestudy}

\begin{tcolorbox}[float=!ht, 
                  width=\linewidth, 
                  colback=yellow!10!white, 
                  colframe=yellow!40!white]
\emph{``Data centers will use 8\% of US power by 2030, compared with 3\% in 2022.''}

\medskip

\hfill -- Goldman Sachs, \citeyear{GoldmanSachs2024}
\end{tcolorbox}

According to \cite{loten2023datacenter} the total data center AI workload consumption in Northern Virginia (NV), VA, the US was 2132~MW in 2023. Thus, the annual data center energy consumption can be estimated as 18.6~million~MWh. Considering the carbon footprint of energy production in NV of 0.3 tonne CO$_2$ per MWh as per \cite{statista2022carbon} the total annual CO$_2$ emissions of NV data centers can be assessed as 5.58~mln tonnes CO$_2$. From the authors of XTAT \cite{Phothilimthana2021AFA} we take 5\% as a reference number for the runtime speed-up across a diverse dataset of 150 neural architectures. Speeding up AI workloads by 5\% with the more efficient execution would reduce CO$_2$ emissions by 275'000 tonnes CO$_2$ yearly in NV alone. This is equivalent to the annual emissions of 36'000 households (approximately 50\% of all NV households). Even though it is yet to be determined how to estimate the real acceleration of computation based on the values of Kendall's $\tau$, we expect the effect to be similar or superior to XTAT \cite{Phothilimthana2021AFA}.

\section{Conclusion}
\label{sec:conclusion}

The proposed novel TGraph neural network architecture establishes a state-of-the-art on the TpuGraphs dataset. A significant contribution to the performance comes from channel-wise self-attention and cross-configuration attention operations. The latter acts as one of the batch normalization techniques, allowing the exchange of information between individual samples, which improves performance in ranking problems.

In general, more efficient ML-based tensor compilation methods have a very positive societal impact. Firstly, they decrease energy consumption and CO$_2$ emissions of data centers, consequently helping to fight climate change. Secondly, they help to free software engineers from the tedious labor of re-implementing lots of highly specialized computational kernels for the constant flow of hardware releases. Even though it may seem that it is a case of "AI taking over people's jobs", in fact, the achieved extreme efficiency of digital infrastructure like data centers may cover the needs of people to the extent that they do not need to work or can opt to dedicate themselves to more human-centered activities.

\section{Limitations}
\label{sec:limitations}

The proposed neural network architecture is limited to predicting the runtimes of a static tensor program that can be represented as a computational graph. Another limitation is that the proposed method is not able to learn the behavior of the tensor program if the behavior is dependent on the values of input or intermediate data. As a machine learning algorithm, the proposed method requires a substantial amount of training data. In the absence of a diverse sample of benchmarked architectures, the domain gap between the training graphs and the unknown test graphs may be big enough, and the model is not able to generalize to it. The proposed method does not provide any guidance on how to choose the graphs for the creation of the training dataset. The proposed method does not generalize to unknown operators. New graphs with the new operator must be added to the training data in order for the model to learn the information about its contribution to the runtime. An ML model trained on one hardware (TPU) does not necessarily generalize to other hardware (GPU, CPU, etc) and must be re-trained for other hardware. Lastly, the proposed solution addresses two compilation sub-problems: tensor layout selection and tensor tiling selection, whereas there are more sub-problems to be solved by tensor compilers.

\ifdefined\isaccepted

\section*{Acknowledgments}
This work was supported by the SDAIA-KAUST Center of Excellence in Data Science and Artificial Intelligence (SDAIA-KAUST AI).

This work was supported by the prizes of the 1st and the 2nd winning places of "Google - Fast or Slow? Predict AI Model Runtime" Kaggle competition.

\fi

\nocite{langley00}

\bibliography{main}
\bibliographystyle{mlsys2025}





\end{document}